# Evaluation of Automated Speech Recognition Systems for Conversational Speech: A Linguistic Perspective

Hannaneh B. Pasandi*, Haniyeh B. Pasandi*

## ABSTRACT

Automatic speech recognition (ASR) meets more informal and free-form input data as voice user interfaces and conversational agents such as the voice assistants such as Alexa, Google Home, etc., gain popularity. Conversational speech is both the most difficult and environmentally relevant sort of data for speech recognition. In this paper, we take a linguistic perspective, and take the French language as a case study toward disambiguation of the French homophones. Our contribution aims to provide more insight into human speech transcription accuracy in conditions to reproduce those of state-of-the-art ASR systems, although in a much focused situation. We investigate a case study involving the most common errors encountered in the automatic transcription of French language.

## 1 INTRODUCTION

There are between 6000 and 8000 different languages spoken in the world. When learning a new language, the distinctions between them appear rather evident because one has to learn a new set of sounds, a new set of words, and a new way to put them together. However, linguists have discovered a plethora of robust patterns that are shared by many languages [39, 72]. These universals of language, which may be devoid of exceptions or significant statistical tendencies instead [25], are essential for explanatory theories of language organization. It has been argued that some aspects of human languages, such as their ability for **productivity** [18] and communicative **effectiveness** [24, 34], are universal. Language is the only animal communication method that takes such a wide variety of forms. The recent [39] study shows an exciting observation using large-scale functional magnetic resonance imaging. (fMRI) about language universality investigation. They tested native speakers of 45 languages across 12 language families (Afro-Asiatic, Austro-Asiatic, Austronesian, Dravidian, Indo-European, Japonic, Koranic, Atlantic-Congo, Sino-Tibetan, Turkic, Uralic, and an isolate, Basque).

The study aimed to determine whether the fundamental characteristics assigned to the "language network" based on data from English and a few other dominant languages also apply to languages with a wide variety of typologies. The participants completed a variety of tasks as well as two naturalistic cognition paradigms to look at correlations in brain activity among the regions supporting high-level cognition connected to language regions and between language regions and regions of another network. Based on the results shown in Figure 4, the left frontal, temporal, and parietal cortex's lateral surfaces are extensively activated.

As mentioned, there is a correlation between a diverse set of languages. One of the important questions to ask is why highly intelligent language models cannot capture such features. Why are ASR systems on the far side of understanding sometimes even simple English sentences?

Based on the aforementioned studies *The goal of this visionary study is to discuss novel research directions to close the current gaps between language misinterpretations via ASR systems. What are the root causes of such issues? What are the wrong or limited assumptions?*

To answer these questions, we present *French language homophones* as our case study to the emergence of this research and its emerging topics. In the following section, we start by defining what homophones are and their role in language.

---

*Authors contributed equally.

## 2 HOMOPHONES
### 2.1 Why homophones?

Ambiguity abounds in human languages. This is most noticeable in homophony when two or more words sound the same yet have different meanings. Why would a system designed for efficient, effective communication be riddled with ambiguity? According to some theories, ambiguity is a design element of human communication systems, allowing languages to recycle their most optimum word forms. Especially those that are short, frequent, and phonetically well-formed) for various meanings, estimates of the rate of homophony in English range from 7.4% to 15% [58].

One of the pioneer works [64] tries to answer why we need homophones. [55, 56, 64–66] Based on their study, the amount of homophony that results from these basic, distributional properties of languages alone is currently unclear. Therefore, neither a case for nor a case against Zipf's proposed direct push for homophony that is driven by efficiency exists. The two main questions in the current work are. First, how much of the homophony observed in actual human lexica can be attributed to unobjectionable, indirect characteristics like length and phonotactic regularities, absent any direct pressure to recycle preexisting word forms? Furthermore, second, how much are these indirect factors to blame for the concentration of homophony in the best lexical areas? To address these questions, they created five sets of synthetic lexica that replicate the phonotactic patterns and word lengths seen in the actual lexica of English, Dutch, German, French, and Japanese. They have proposed a mechanism: rather than concentrating certain sequences in a single word form, they might "smooth out" high-probability phonotactic sequences throughout lexical areas. The result is lexica which may be slightly less ideal in phonotactic terms but may better meet other needs of the users. We refer the interested reader for more details about their work.

Typically, it is believed that demands for efficient, effective communication influence human languages. However, ambiguity raises the chance of misinterpretation while also increasing the effort required to comprehend [24]. This is clarified by comparing human and programming languages. Programming languages often tolerate no ambiguity at all, as they were created for effective and error-free communication. Hence why do human languages persist in encoding various signals in the same way? Why is homophone usage so prevalent?

Homophones and minimal pairs pose serious difficulties for automatic speech recognition [23]. In sentence-level speech recognition, the process of homophone disambiguation is done initially through the selection of a homophone word that has a dominating meaning than its other homophone(s) [8, 40], or within a stochastic language model [35].

### 2.2 Homophones in Voice Anatomy Diagram

As shown in Figure 1, words (sounds) in general, including homophones, are usually generated from different parts of the voice anatomy diagram, and their utterance is either in the upper or lower part of the Larynx depending on the spoken language. For example, in the English language, the utterance may come from the upper part of the Larynx as opposed to the French language. This by itself could cause various features even if the same language is spoken by an individual.

There have been numerous studies in past years that investigate this topic and provide innovative solutions [9, 42]. However, less attention has been paid to the homophone issues in ASR systems. Authors in proceed with a different perspective to address the homophone issues by *automatically* locating and distinguishing homophones in code-switched data to help code-switched voice recognition. Maas et al. [38] presented an approach to speech recognition that used a neural network to map acoustic input to characters and a beam search decoding procedure by Lexicon-Free Speech Recognition (LFSR). Another research [31] showed that the speech recognition technique used in Apple's iOS (For Siri) misinterprets the user input and picks the wrong homophones spoken in a noisy or noise-free area in Word Level Speech Recognition. There are different research directions to address the challenges of language ambiguity using different approaches [1, 4, 13, 16, 28, 57] and different languages [14, 68].

Some studies focused on the acoustic analysis of homophone words. Clopper et al. [20] demonstrate how a phonetic model can be used to automatically disambiguate homophones using the example of German pronouns in short or syntactically incomplete contexts. The model is trained based on Deep Neural Networks (DNNs), and the results show that homophones can be disambiguated reasonably well using acoustic features. According to research in [20], both the homophones with dominant and secondary meanings are selected through a supervised method in the syntactic model. The recent research by Gosh et al. [23], also studies the speech recognition techniques in word-level speech recognition and suggests unsupervised reinforcement learning based Artificial Immune Algorithm to reduce the homophone ambiguity from any existing speech-to-text techniques in word-level speech recognition. The comparison of these different studies highlights the importance of lexical context for accurate human transcription.

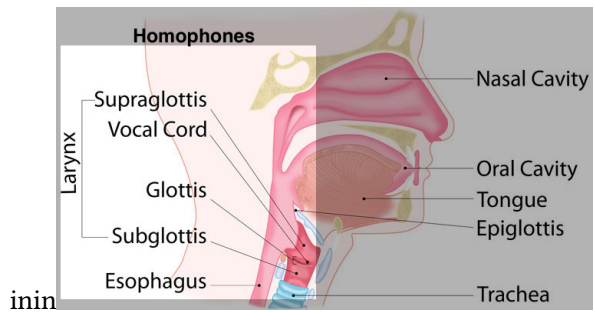

Figure 1: Human Voice Anatomy Diagram and Homophones.

In line with [23], our contribution aims to provide more insight into human speech transcription accuracy in conditions to reproduce those of state-of-the-art ASR systems, although in a much more focused situation. We investigate a case study involving the most common errors encountered in the automatic transcription of French: the erroneous transcription of homophonic words "leur"/"leurs" [1]. (Possessive Pronouns) and "leur" (Object Pronouns). By focusing on this very particular case, we raise the question of whether humans use quantifiable information for such homophone disambiguation that has not been exploited by ASR systems so far, focusing on acoustic differences of homophone words.

The studied items "leur", "leurs", and "leur" are frequently used in a variety of contexts, both written and spoken. However, these words correspond to different parts of speech and occupy distinct positions within sentences. In French, the syntactic differences between these words are often accompanied by acoustic and prosodic peculiarities which might help humans to distinguish them[2]. When it comes to voice assistants such as Siri, it can misunderstand the word in clever ways even if the person talks loudly and in a quiet environment due to selecting an incorrect homophone or just mishearing the individual and repeating phrases that only slightly resemble what they have said. Even if the person attempts to correct it while in the same session, they can continue using the wrong information. As a single example, this shows the importance of the disambiguation of homophones in different languages.

In the French language case, we categorize the issues that these homophone words create in two parts.

Firstly, when these words have different syntactic positions in a sentence, they tend to be recognized easier regardless of their identical spelling.

**Example One.** On leur a dit: Object Pronoun (They were told) Leurs parents ne sont pas là: Possessive Pronouns (Their parents are not there).

Secondly, when these words are used as Possessive Pronouns in both singular and plural (leur: singular, leurs: plural), the disambiguation of homophones becomes challenging as the plural "s" is not pronounced in French unless it is followed by a word starting with a vowel. See the below examples:

**Example Two.** Leur livre (Their book) Leurs livres (Their Books)

Please note that the two sentences are pronounced precisely in the same way.

There are other homophones disambiguations, as well. Such disambiguation of these homophones relies on the **sentence-level** context. However, our experiments will show that our two categorizations are not enough.

---

[1] Other examples include the following monemes in French: sain, saint, sein, ceint.

[2] In contrast to the French language, for example in Spanish the "s" is pronounced, which eliminates the inaccuracy rate of words ASR systems. Examples: su libro (his/her book), sus libros (their books).

Table 1: Results of Siri transcriptions

| 30 Sentences transcribed by Siri | Types of errors |
|---|---|
| 10 corrects | leur/leurs correctly transcribed by the system |
| 17 confusions | 17 leur/leurs confusions |
| 3 other errors | Deletion, transcription of target word alone or within a syntagm. |

## 3 PROPOSED SOLUTION: HOMOPHONE IN ASR SYSTEMS

Machine learning approaches have been successfully applied from communication [5, 6, 45, 46, 49–52] to speech [17] to video [47, 48] to text [2, 29]. Task-agnostic text representation learning has received much interest recently in the NLP field due to its outstanding performance on many downstream tasks ( [21, 60, 63] such as speech recognition. Unsupervised speech representation learning has also recently successfully enhanced various speech-related tasks [37].

As shown in Figure 3, the current speech recognition pipeline has five steps, as illustrated. One proposed approach is to have a multimodal ASR pipeline to embed the homophones as texts inspired by cite-modal3,modal2, modal1.

## 4 DATA PROCESSING AND EXPERIMENT SETUP

To perform our experiments, we used The SIWIS French Speech Synthesis Database [3] on Apple Siri since it is one of the most accessible and the most used ASR system. We selected 30 recordings with the aforementioned homophones to Apple Siri. In the second phase, we transcribed another 30 sentences that were read by Apple Siri. The experiment aimed to check if a human transcriber can correctly identify the homophone mentioned above words in limited contexts, which have proved to be ambiguous for the ASR system. Finally, the acoustic differences of these homophone words were studied.

[3]https://datashare.ed.ac.uk/handle/10283/2353

### 4.1 Apple Siri transcription of homophones

Table 2 demonstrates some common examples of transcription errors of our selected homophone words. All of the sentences have around 4 to 7 N-grams.

Table 2: Apple Siri transcription of homophones

| Reference transcriptions | Transcription of Siri |
|---|---|
| On leur a parlé. | On aura parlé. |
| On leur a donné leurs livres. | On leur a donné leur livre. |

### 4.2 Human transcription of homophones

For 10 sentences, we performed a local language model (LM) condition test on the 30 sentences focusing here on "leur" vs "leurs" confusion. After listening to each sentence, we had to fill in a written version of the 30 sentences using the most accurate item "leur" or "leurs". The rationale of this test is two-fold: syntactic/semantic information of the written sequence contributes to solving ambiguity; **humans explicitly focus on local ambiguity.**

| ASR System | Human |
|---|---|
| 10 Corrections | 25 Corrects |
| 17 confusions | 5 Confusions |
| 3 other errors | N/A |

Table 3: ASR Vs. Human error rates

Evaluation of Automated Speech Recognition Systems for Conversational Speech:
A Linguistic Perspective

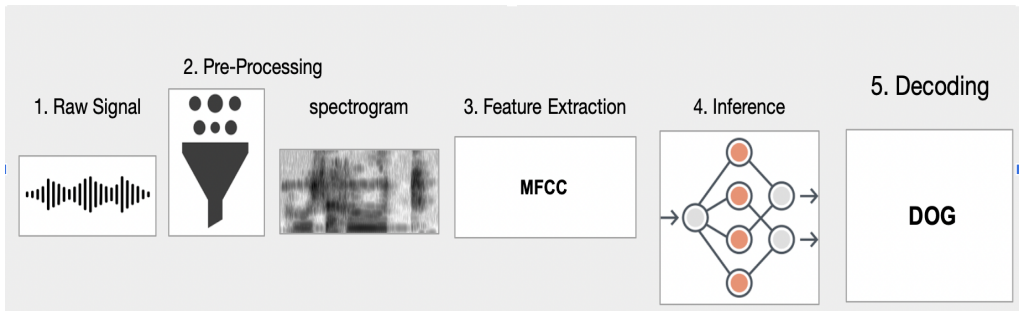

Figure 2: ASR Pipeline: The processes that compose the pipeline for automatic speech recognition. 1) A microphone is used to record the audio. 2) A low pass filter is used beforehand to eliminate any basic noise or high frequencies. 3) The most crucial aspects of the audio sample are extracted by running the audio through a feature extraction function. 4) The features are given to the model for inference, which generates a transcription that humans can read.

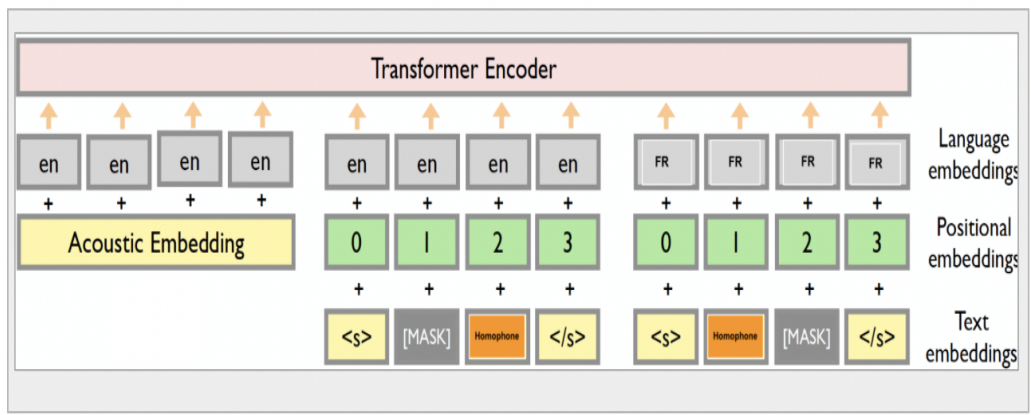

Figure 3: Potential multimodal ASR Pipeline.

## 5 RESULTS

To evaluate, human error rates are compared to ASR word errors in terms of reference transcriptions. As a general observation, human errors were considerably less than ASR systematic errors. Differences in ratings for similar ambiguous syntactic structures suggest that word-level recognition may help in operating the right choice in terms of target word selection using an Artificial Immune System Algorithm in Word Level Speech Recognition.

When looking at different types of errors for each of the target words, one may notice that the more ambiguous the local context, the more frequently the correct solution is missed by human listeners. In particular, participants had difficulty distinguishing between the singular and plural form of "leur" when it was followed by a word that the first letter was a consonant, i.e. "leur" livre (their book) or "leurs" livres (their books) as they are similarly pronounced and have the same syntactic structure. This finding suggests that this type of homophone ambiguity is problematic for both the ASR system and humans.

The most popular benchmarking techniques for ASR systems are seldom applicable to conversations. Word error rate (WER), a common statistic, compares In terms of insertions, deletions, and substitutions, ASR output is compared to reference transcripts. Although helpful, it is not without drawbacks. One drawback is that it prioritizes insertions above deletions. Additionally, it ignores the fact that words come in various forms, despite research on ASR transcription errors in English

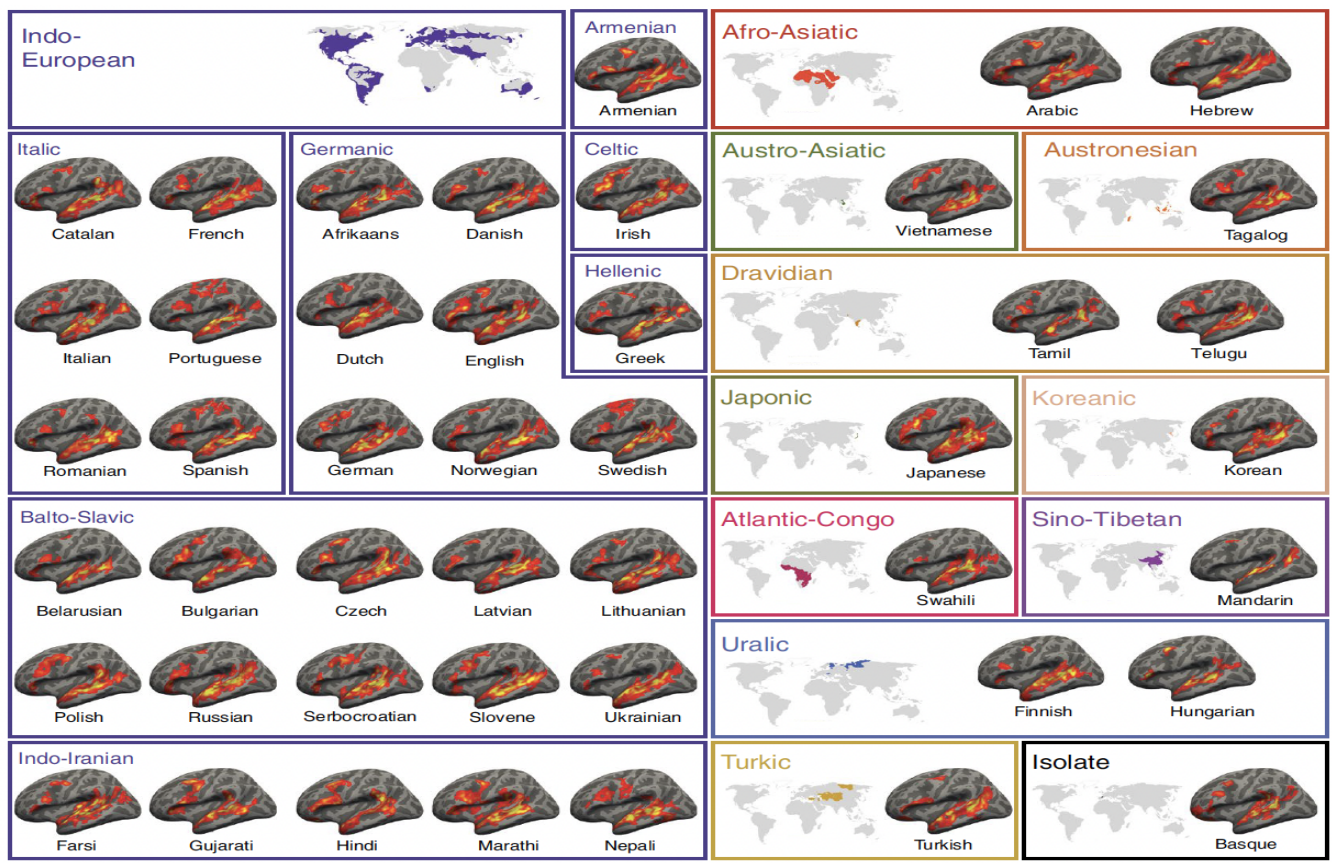

Figure 4: Native speakers of diverse languages form a language network. Activation maps in the LH of a sample participant for each language in the Alice language localizer comparison (Native-language > Degraded-language. The general structure of the language network is similar across 45 languages, and the observed diversity is consistent with that previously documented for speakers of the same language. FreeSurfer created a significance map for each participant, smoothed using a Gaussian kernel with a 4-mm FWHM and thresholded at the 70th percentile of the subject's positive contrast. FreeView produced the surface overlays on the average template's 80% inflated white-gray matter boundary. The 80th and 99th percentiles of each participant's positive-contrast activation are shown by opaque red and yellow, respectively. (These maps were just used for visualization [39]. There are more analyses and references given in the figure description. We encourage the interested user to refer to [39].

demonstrating that these errors are more likely to occur for conversational interjections [70].

To solve such issues, there is a massive effort that needs to be conducted. The authors suggested that each natural language processing (NLP) should be user-centric. On-device personalization of speech models is a research topic that is currently in progress [15]. Therefore, we need more customized language models that fit the language context in its cultural and societal settings. Based on the study in [71], user modeling and NLP are closely entangled. Several aspects of NLP, such as analyzing users' words and correcting potential misunderstandings, can be aided by user models.

NLP models are not one-size-fits-all. Therefore, the authors believe that to improve NLP models, the user models should be personalized and customized for diverse users. Such a challenge may require ASR systems (voice/virtual) assistants to deploy on-device models that will continuously learn how to interact with the user and consequently improve. Although companies like Apple have already started offline training on devices with Siri activated on them. Another important research direction for such models would be not only to improve the limited language assumptions they have but also to keep users' privacy and AI ethics in mind while building such models.

# 6 CHALLENGES

Languages used by humans change over time to improve communication. However, efficiency entails compromises: what works well for speakers may not always work well for listeners. How can different languages balance these conflicting demands? We concentrate on Zipf's law of meaning-frequency, which states that common word forms have more meanings. One way to look at this regulation is as a speaker-oriented pressure to recycle common word forms. However, there are still hundreds of different word forms in human languages, indicating a balancing, comprehender-oriented demand. There is an evolutionary pressure toward efficiency.

Explaining language structure, both methodological and empirical is challenging.[4].

## 6.1 Specification and measurement of complexity

Efficiency can be measured using information-theoretic ideas, but there is less consensus on how to describe a language's complexity or learnability. The length of a language's description in a meta-language, such as [10, 19, 27, 32, 33, 41, 53, 59, 67, 69], can be used to quantify a language's complexity, in particular; however, this measurement depends on the description meta-language that is being utilized. There is still a possibility for significant magnitude disparities even though this complexity assessment is partially independent of the meta-language [36]. In order to further describe and make precise the relationship between learnability, compositionality, systematicity, and iconicity, it will be necessary to clearly establish the appropriate complexity measure. In order to show that a specific language component is efficient in terms of communication and learnability, it is frequently necessary to compare the observed language data with some counterfactual baseline reflecting what a language may look like without pressure for efficiency.

For instance, if we wanted to say that a certain phrase is the most effective method to convey a specific message in a given context, we would need to compare it to all the other possible things a speaker could have said given all the degrees of freedom accessible to them, including pragmatics. Such baseline utterance generation would be incredibly difficult. There is currently no mechanism for creating baseline utterances and languages that is universal and effective in all situations.

## 6.2 Information-theoretic quantity measurement

To quantify communicative effectiveness, it is necessary to measure the information content, entropy, etc., of words, phrases, and messages. By fitting probability distributions drawn from corpora, these values are calculated. Without very big datasets, it is challenging to obtain accurate estimations of these information-theoretic properties [11, 44]. Use Google's web-scale n-gram corpora as an illustration [54]. Empirical studies of many languages are challenging due to the dependency on huge datasets, and many researches are limited to modern English because it has the greatest data.

## 6.3 Specification of communicative utility

Efficiency in the context of information theory refers to how well a language conveys arbitrary signals. However, in actual natural language conversation, additional elements play a role in how useful utterances are. Languages, for instance, are sometimes ingrained with taboo terms that must be avoided lest the speaker suffers societal penalties. The avoidance of such terms must be taken into account when determining the utility of a statement; even if a statement is otherwise effective, it will still be avoided if it contains a taboo word. Whether these factors may be stated as a component of the general communicative efficiency as described above or whether they call for the inclusion of new components to the equations characterizing communicative utility is still unknown.

---

[4]This section is inspired by the communication efficiency from [24]

## 6.4 Population and Cultural Differences

The sound repertoires, grammatical structures, and lexical organization of languages differ around the globe. The people where these languages are spoken also differ greatly in terms of social structure, access to technology, and cultural values. These latter distinctions have the potential to alter how communicative demands are distributed: in some populations, it will frequently be more crucial to express a specific form of meaning than in others. New words to name and characterize new technology appear almost soon, for example. In some situations, these distinctions in communicative demands may be mirrored in specialized grammaticalized subsystems within the language, like the grammar of honorific expressions in Japanese [62].

The quantity of common knowledge shared among speakers may vary depending on factors including population structure [22] and the prevalence of non-native speakers [12], which may therefore influence the linguistic encodings of intended speaker meanings necessary to enable accurate communication. Uncertainty persists over the extent to which these factors can result in explanatory theories of cross-linguistic variations and historical change's routes. Language users and learners face a trade-off between efficiency and complexity, which can be effectively resolved by the variety of ways that languages can be arranged around the world.

Although major linguistic characteristics can change quickly, not all changes are equally likely. In early English, for instance, the favored subject, object, and verb order totally changed over a few hundred years [165], but not all changes among the six logically feasible orderings are equally plausible. An explanatory theory of the changes in the linguistic structure must incorporate historical language change models, but our knowledge of these models' limitations is still restricted [26, 30, 43].

## 7 CONCLUSION AND FUTURE WORK

The main ecosystem of natural language use is conversation. ASR systems are increasingly exposed to conversational contexts [7]. They cannot, however, manage naturally occurring dialogues. In this paper, our contribution provided more insight into human speech transcription accuracy in conditions to reproduce those of state-of-the-art ASR systems, although in a much-focused situation. We studied a specific case of homophones in French and how ASR systems fail to correctly perform specific tasks. We would like to conclude by asking how far we are from pure voice-driven human language technologies.

In our future research, we will examine if differentiating between homophones would be improved by adding a language model (n-gram or transformer based). Comparing character-level, word-level, and sentence-level models for identifying homophones is another aspect to take into account.

Another interesting direction would be to study the human speech ambiguity interaction with ASR systems. While current ASR systems claim to handle spontaneous speech, this frequently takes the form of voice commands that, with some exceptions, do not require a rapid verbal response [61]. It has been found that people utilize more tightly controlled, fluent, and constrained speech while utilizing voice instructions than when speaking to another human or device over the time of aging or gesture production, and dysfluency in Speech [3].

Evaluation of Automated Speech Recognition Systems for Conversational Speech:
A Linguistic Perspective

[72] . . Litres, 2022.